\documentclass{article}

\usepackage[margin=1in]{geometry}
\usepackage{authblk}
\usepackage[numbers]{natbib}
\usepackage{amsmath,amssymb}
\usepackage{graphicx}
\usepackage{subcaption}
\usepackage{booktabs}
\usepackage{multirow}
\usepackage{placeins}

\title{Complex Equation Learner: Rational Symbolic Regression with Gradient Descent in Complex Domain}

\author[1]{Sergei Garmaev}
\author[2]{Maurice Gauché}
\author[1]{Olga Fink}

\affil[1]{Intelligent Maintenance and Operations Systems Laboratory, EPFL, Lausanne, Switzerland}
\affil[2]{Mathematics Section, EPFL, Lausanne, Switzerland}

\date{}

\begin{document}

\maketitle

\begin{abstract}
Symbolic regression aims to discover interpretable equations from data, yet modern gradient-based methods fail for operators that introduce singularities or domain constraints, including division, logarithms, and square roots. As a result, Equation Learner-type models typically avoid these operators or impose restrictions, e.g. constraining denominators to prevent poles, which narrows the hypothesis class. We propose a complex weight extension of the Equation Learner that mitigates real-valued optimization pathologies by allowing optimization trajectories to bypass real-axis degeneracies. The proposed approach converges stably even when the target expression has real-domain poles, and it enables unconstrained use of operations such as logarithm and square root. We Validate the method on symbolic regression benchmarks and show it can recover singular behavior from experimental frequency response data.
\end{abstract}


\section{Introduction} \label{sec:introduction}
Symbolic regression (SR) and model discovery aim to identify concise, human-interpretable mathematical expressions that govern observed data.
In contrast to standard regression methods, which estimate parameters within a fixed functional form, SR jointly searches  over both the structure and parameters of candidate models, enabling the direct recovery of explicit analytical relationships from data.
This capability is particularly valuable in scientific and engineering domains, such as fundamental and applied physics, cosmology, neuroscience and biology, and complex engineering systems, where interpretable models are essential for extracting physical insight,  revealing the underlying mechanisms that govern system behavior, and connecting data-driven discoveries to established theoretical principles.
As a result, SR has become a central tool in data-driven scientific discovery, with applications ranging from physics and chemistry to systems biology and materials science. 

Symbolic regression methods can be broadly grouped into implicit learned model-based and explicit search-based approaches. Recent years have seen substantial progress in deep learning-based SR, including transformer models \cite{biggio2021neural} and reinforcement learning methods \cite{petersen2021deep, tian2025interactive}. However, these approaches remain intrinsically implicit, i.e. they learn an inverse mapping from data to equations tied to the training distribution of expressions, domain, and sampling schemes. Consequently, their behavior under distribution shift cannot be systematically controlled especially in extrapolative regimes. In contrast, explicit search-based methods do not rely on such learned priors and are typically not sensitive to the expression structure changes compared to the training setup. Among these, genetic programming (GP) has historically been a dominant paradigm for symbolic regression \cite{makke2024interpretable}. However, GP remains inherently heuristic and combinatorial, limiting their applicability in high-dimensional settings due to the combinatorial expansion of the expression space.

Another class of SR methods formulates model discovery as a gradient-based optimization problem by minimizing data approximation error \cite{makke2024interpretable}. Some approaches reduce SR to linear regression over a fixed  library of functions, such as Sparce Identification of Nonlinear Dynamics (SINDy) \cite{brunton2016discovering}, which selects a sparse linear combination of of candidate library terms. SINDy performs well when  true dynamics are sparse in the chosen basis, but degrades when essential functional forms are absent from the library and becomes less scalable as the library growths \cite{brunton2016discovering}. More expressive non-linear models, such as Kolmogorov-Arnold Networks (KANs), learn univariate edge functions that can yield interpretable components but require post-processing to obtain compact closed-form equations \cite{liu2025kan}. The Equation Learner (EQL) instead builds hypotheses directly from analytic operators in a differentiable architecture, enabling extraction of explicit symbolic expressions after training \cite{martius2016extrapolation}. Extensions incorporating modified division operators increase EQL’s expressivity \cite{sahoo2018learning}, yet they enforce a fixed sign on the denominator, substantially restricting the class of representable rational forms.

Among gradient-based approaches, the EQL is seen as the most promising approach, as its architecture explicitly composes operators within a differentiable network. By construction, this model contains all possible symbolic expressions up to a predefined compositional depth, while the learning procedure reduces to optimizing internal weights, enabling scalable discovery of closed-form expressions. However, in its current form, EQL is limited to a relatively narrow set of operations. Fundamental operators such as division, logarithms, and square root are difficult to incorporate reliably due to intrinsic optimization pathologies, as analyzed in Section~\ref{sec:preliminaries}. These operators are ubiquitous in physical laws and constitutive models, and their exclusion  substantially restricts the scope of symbolic model discovery. This limitation motivates the need for a principled symbolic regression framework that preserves explicit, interpretable structure while supporting stable optimization over a substantially broader class of nonlinear operators, including those that may induce singularities.

To address these gaps, we propose  the Complex Equation Learner (CEQL), an extension of EQL that enables stable gradient-based learning of rational expressions involving division. CEQL relaxes real-valued training by allowing network weights to evolve in the complex domain while projecting model outputs back onto the real axis. Optimization in the complex space allows  training trajectories to bypass real-axis degeneracies that, in purely real-valued optimization, lead to sign-changing gradient cancellations. As a result, CEQL enables the reliable learning of explicit denominators and near-singular structures using gradient descent. We demonstrate that CEQL  identifies poles in rational target expressions and recovers interpretable pole-bearing symbolic forms. In particular, CEQL discovers rational models that capture resonant behavior in frequency response functions, where inspection of the learned denominator directly relates pole locations to changes in system parameters. Moreover,  operating in the complex space removes the standard real-valued domain restrictions of operators such as logarithms and square roots, allowing them to be incorporated  into a differentiable symbolic architecture without ad hoc constraints. Finally, because CEQL performs gradient-based optimization over a fixed operator library rather than a combinatorial tree search, it scales more favorably with input dimensionality than GP-type methods.

\section{Preliminaries} \label{sec:preliminaries}
\subsection{Gradient Pathologies in Real-Valued Optimization} \label{sec:gradient_pathologies}
Consider the problem of approximating the function $f(x) = 1/x$ on the interval $[-\ell, \ell]$. Let $\{x_i\}_{i=1}^N$ be a set of sample points drawn from this interval, and let the parametric model be $\hat{f}(x) = 1 / (x + a)$, where $a \in \mathbb{R}$.  We formulate the approximation task as the optimization problem
\[
a^\star = \arg\min_{a \in \mathbb{R}} \; \mathcal{L}(a),
\quad
\mathcal{L}(a)
= \frac{1}{N} \sum_{i=1}^N 
    \left( f(x_i) - \hat{f}(x_i) \right)^2 .
\]
Here, $a^\star$ denotes the minimizer of the loss $\mathcal{L}(a)$, which in this setting is attained at $a^\star = 0$. Despite its apparent simplicity, this problem exhibits fundamental obstacles for gradient-based optimization methods. To see this, consider the gradient of  $\mathcal{L}$ with respect to $a$:
\begin{equation} \label{eq:sign_changing_gradient}
\begin{aligned}
    \frac{\partial \mathcal{L}}{\partial a}
    &= \frac{\partial}{\partial a}
       \left( \frac{1}{N} \sum_{i=1}^N 
            \left( \frac{1}{x_i} - \frac{1}{x_i + a} \right)^2 \right) \\
    &= \frac{2}{N} \sum_{i=1}^N 
        \left( \frac{1}{x_i} - \frac{1}{x_i + a} \right)
        \frac{1}{(x_i + a)^2}.
\end{aligned}
\end{equation}
Let
\[
g(x_i, a)
= \left( \frac{1}{x_i} - \frac{1}{x_i + a} \right)
  \frac{1}{(x_i + a)^2}
\]
denote the per-sample contribution to the gradient. The behavior of the total gradient is governed by this term, which simplifies to $g(x_i,a)=a\left(x_i (x_i+a)^3\right)^{-1}$. For any fixed  $a$, the sign of $g(x_i,a)$ is therefore determined by the signs of $x_i$ and $x_i + a$. Consequently, the per-sample gradient contribution $g(x_i,a)$ changes sign at $x_i = 0$ and $x_i = -a$, producing opposite gradient contributions from different intervals of the domain. When samples lie on both sides of these sign-change boundaries, the terms $g(x_i,a)$ in the gradient sum cancel, leading to oscillatory or near-zero gradients. As a result, the total gradient does not, in general, provide a consistent descent direction toward the minimizer $a^\star = 0$. The same sign-changing mechanism arises whenever denominators introduce additional terms, as each term induces further sign-changes in the gradient.

This observation shows that the difficulty encountered above is not a consequence of poles in the model, nor of isolated pathological points in the domain. It originates from the intrinsic sign-changing structure of the gradient in the real-valued parameter space, which leads to substantial cancellations across the data and results in a flat loss landscape. An analogous effect arises for the gradient when approximating $f = \sin(x)$ with the parametric model $\hat{f}(x)=\sin(bx)$, one can derive the gradient of the loss w.r.t. parameter $b$, and verify that the  per-sample gradient contributions similarly change sign across the domain, leading to the similar cancellation mechanism. Thus, the main challenge faced by gradient-based methods is a structural property of the function’s behavior in the real-domain, rather than an artifact of poles.

\subsection{Multi-valued Functions and Domain Restrictions}
In addition to the gradient pathologies discussed above, certain operators introduce a distinct class of difficulties related to domain restrictions induced by the branch cuts of multi-valued functions. A representative example is the logarithm. In the real domain, the natural logarithm $\ln(\cdot)$ is defined only for strictly positive arguments. When implemented within a real-valued EQL network, the input to the logarithm corresponds to  an intermediate activation produced by preceding layers and therefore varies across samples during training. As a consequence, some samples may drive this activation outside the domain of definition, leading to undefined forward evaluations whenever the activation is negative or approaches zero.

From a complex-analytic perspective, this restricted real-domain definition reflects the fact that the logarithm is multi-valued in the complex plane. For a complex argument $z \in \mathbb{C}\setminus\{0\}$, the logarithm admits the family of values
\[
    \ln z = \ln|z| + i(\arg z + 2\pi k), \qquad k \in \mathbb{Z}.
\]
Any single-valued realization of $\ln(\cdot)$ therefore requires the selection of a branch, which necessarily introduces a branch cut along which the argument $\arg z$ is discontinuous. The domain restriction of the real logarithm can thus be interpreted as a consequence of this branch selection, rather than as an intrinsic property of the logarithm itself.

The same considerations apply to the square root operation. The square root $\sqrt{x}$ is commonly expressed using the identity
\[
    \sqrt{x} = x^{\frac{1}{2}} = \exp\!\bigl(\frac{1}{2} \ln x\bigr),
\]
with $x \in \mathbb{R}$. This representation makes explicit that expressions involving $\sqrt{x}$ inherit the same branch-cut-related issues as the logarithm. In particular, unless the base $x$ is constrained to remain positive, the expression becomes ill-defined in the real domain.

A practical workaround, similar in spirit to sign-restricted division used in previous EQL extensions \cite{sahoo2018learning}, is to assume that the argument of the logarithm does not change sign and to replace $\ln(x)$ with $\ln|x|$, and correspondingly $\sqrt{x}$ with $\exp\!\bigl(\frac{1}{2} \ln|x|\bigr)$. While this modification helps to avoid undefined regions in the real-valued formulation, it imposes a strong structural assumption. In realistic settings, the true target expression may change sign, with the logarithm being well-defined only on subsets of the domain actually covered by the data. Enforcing a globally non-sign-changing behavior in the symbolic network therefore prevents the recovery of such expressions, even when all observed samples lie within regions where the target expression itself is well-defined.

\section{Methods}
\subsection{Surrogate Symbolic Operations} \label{sec:surrogates}
To address the challenges described in Section~\ref{sec:preliminaries}, we propose to modify the backbone of the EQL framework by allowing its parameters to take complex values. This complex parametrization enlarges the optimization domain from $\mathbb{R}^p$ to $\mathbb{C}^p$, thereby altering the geometry of the loss landscape and mitigating gradient pathologies that arise in purely real-valued optimization. Model outputs are subsequently projected onto the real axis, and the loss is computed using the real part of the network output.

\textbf{For division operation}, complex-valued parameters introduce a complex shift in the denominator. In the example of approximating $f(x) = 1/x$ with a model of the form $\hat{f}(x) = 1 / (x+a)$, allowing $a \in \mathbb{C}$ smooths the optimization trajectory by avoiding gradient cancellation effects present in real-valued optimization. An illustrative optimization example is given in Appendix~\ref{si:complex_optimization}.

Gradient cancellations arise primarily for parameters appearing in denominators and  do not occur for parameters in numerators. Accordingly, we define a surrogate division operation $b_{\text{div}}$ as
\begin{equation}
    b_{\text{div}} (x, y) = \Re(\Re(x) / y),
\end{equation}
where $x, y \in \mathbb{C}$ are the inputs to the surrogate operator. This design preserves a real-valued symbolic output while retaining extra degree of freedom in the denominator, thereby mitigating the gradient-cancellation effects encountered in real-valued optimization.

\textbf{The logarithm and square root} do not pose intrinsic difficulties for gradient-based optimization within their domains, as they are smooth on $\mathbb{R}_{>0}$. The practical issue instead arises from domain restrictions. In EQL, intermediate activations are unconstrained during training, so the arguments of $\ln(\cdot)$ and $\sqrt{\cdot}$ can become non-positive or approach zero, leading to undefined evaluations on the real axis. This limitation motivated previous EQL variants to replace these operators with “safe” real-valued surrogates such as $\ln(|\cdot|)$ and $\sqrt{|\cdot|}$, at the cost of changing the target function class. In this work, we use the complex logarithm and complex square root, which are well-defined on $\mathbb{C}\setminus(-\infty,0]$, once a branch is chosen. This yields valid forward evaluations and well-behaved gradients almost everywhere, except across the corresponding branch cuts. As discussed in Section~\ref{sec:optimization}, we add a small penalty on the imaginary part during training to keep intermediate values close to the principal branch, while optimizing the real parts of the parameters.

All other unary operators $u_i$ are implemented via real-projected surrogates $u_i^*(x)$ defined as
\begin{equation}
    u_i^*(x) = u_i(\Re(x)) + 0i,
\end{equation}
where $x \in \mathbb{C}$ denotes the complex-valued input activation. That is, the original real-valued operator is applied to the real part of the activation, and  the result is embedded back into $\mathbb{C}$ with zero imaginary component.

Similarly, for binary operators $b_i$, i.e.,  operators with two arguments, we define the surrogate $b_i^*$ as
\begin{equation}
    b_j^* (x, y) = b_j (\Re(x), \Re(y)) + 0i,
\end{equation}
with $x, y \in \mathbb{C}$. This construction is used for all binary operators except division, which is handled by the dedicated surrogate $b_{\text{div}}$ defined above.

Overall, these surrogate definitions enforce a real-valued symbolic hypothesis class at the level of operator evaluations, while still allowing complex-valued parameters and intermediate representations to introduce additional degrees of freedom during optimization for selected operators.

\subsection{Complex Equation Learner}
\begin{figure}[t]
    \centering
    \includegraphics[width=0.7\linewidth]{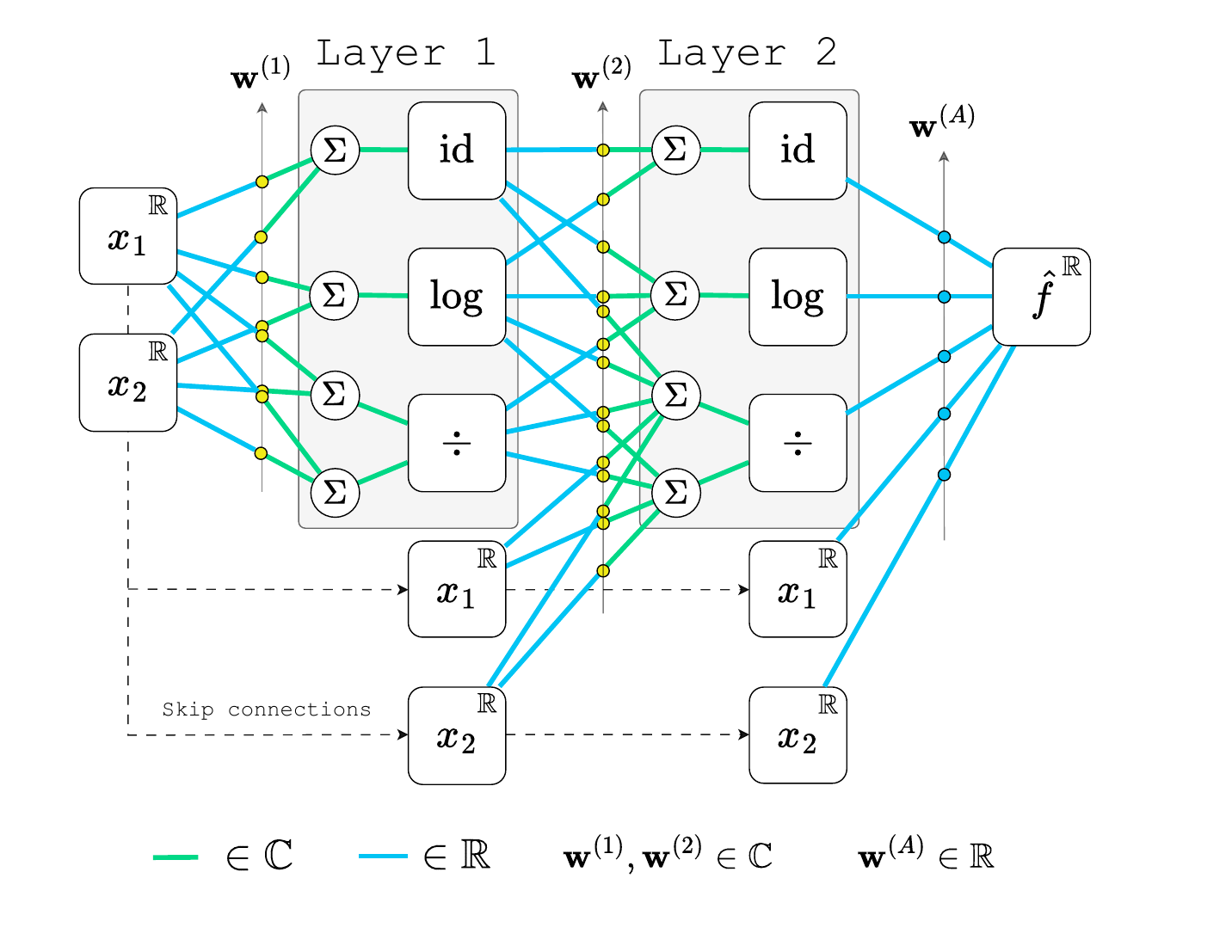}
    \caption{Complex Equation Learner (CEQL) architecture. Internal weights are complex-valued, enabling optimization to bypass real-valued degeneracies induced by division and multivaled operators. While the internal weights are optimized in $\mathbb{C}$, the output of CEQL is projected to $\mathbb{R}$ to minimize discrepancy with the real-valued target values.}
    \label{fig:ceql}
\end{figure}

The proposed CEQL approach builds on the original EQL framework. As a fully connected neural network, CEQL can be viewed as a directed acyclic graph, in which nodes implement symbolic operations and edges carry learnable parameters. In contrast to EQL, CEQL uses complex-valued weights, which enlarge the optimization space and address the division-related gradient issues discussed in Section~\ref{sec:gradient_pathologies}. The graph contains one input node per input variable of the target function and a single output node representing the CEQL prediction, as illustrated in Figure~\ref{fig:ceql}. While internal signals within CEQL are complex, both the inputs and the target outputs are strictly real-valued in all experiments.

Each layer contains a fixed set of operators, referred to as the operator library. In this work, the library consists of unary surrogates $u_j^*$ and binary surrogates $b_k^*$ defined in Section~\ref{sec:surrogates}. The inputs to the operators in layer $L$ are formed by summation nodes,
\[
z^{(L)}_j=\sum_{i=1}^{d_{L-1}} w^{(L)}_{ij}\,h^{(L-1)}_i,\qquad j=1,\dots,m+2n,
\]
where $h^{(L-1)}_i$ denotes the $i$-th activation of layer $L-1$, $d_{L-1}$ is the number of activations in that layer, and $w^{(L)}_{ij}$ is the (complex-valued) weight on the edge from activation $i$ to summation node $j$. The first $m$ summation nodes are passed through unary operators, $h^{(L)}_j=u_j^*(z^{(L)}_j)$ for $j=1,\dots,m$. The remaining $2n$ summation nodes are grouped into $n$ consecutive pairs and passed to the binary operators, $h^{(L)}_{m+k}=b_k^*(z^{(L)}_{m+2k-1},\,z^{(L)}_{m+2k})$ for $k=1,\dots,n$. The model output  $\hat f$ is obtained as a weighted sum of the final-layer activations.

Beyond the original EQL formulation, CEQL incorporates skip connections that route the original input variables directly to deeper layers by concatenating them with intermediate activations. This design is inspired by residual and feature-concatenation connectivity patterns in deep networks \cite{he2016deep}. Skip connections improve gradient propagation when the target expression is shallow and allow the optimizer to select shorter computational paths, thereby simplifying the search over expression structures.

Overall, the resulting architecture spans the space of symbolic expression trees generated by the chosen library up to a prescribed depth. CEQL is trained to minimize the discrepancy between the predicted output and the target values while enforcing sparsity. The sparsity mechanism and optimization procedure are described in Section~\ref{sec:optimization}.

\subsection{Optimization Strategy} \label{sec:optimization}

CEQL is trained by minimizing mean squared error (MSE) between the target values and the real part of the CEQL output, while internal parameters remain complex-valued. The $\ell_2$ objective is augmented with auxiliary terms that promote sparsity and stabilize optimization. Training is organized into three phases.

In the first phase, we minimize the training MSE augmented with a sparsity-inducing penalty on the weights and a penalty on their imaginary components. The sparsity term encourages small-magnitude connections to shrink toward zero, while the imaginary penalty biases parameters toward the real axis, keeping intermediate evaluations of multi-valued operators close to the principal branch. The corresponding regularization coefficients are chosen several orders of magnitude smaller than the MSE term. At the end of this phase, all connections with magnitude below a fixed threshold are pruned and subsequently held at zero. We then perform a cascade cleanup to remove edges and nodes that are no longer connected to the output.

In the second phase, training continues on the pruned model using a stronger sparsity penalty and the same imaginary-weight penalty. To further simplify the expression and accelerate convergence, we perform iterative pruning cycles during training. At each cycle, a prescribed fraction of connections with the smallest impact on the batch loss is removed, while enforcing a minimum number of remaining edges to control the maximum size of the resulting expression tree. After each pruning step, disconnected branches are again removed via cascade cleanup.

In the third phase, sparsity regularization is disabled and the remaining parameters are fine-tuned using only the MSE loss and the small imaginary-weight penalty. This final refinement improves numerical accuracy while preserving the discovered symbolic structure. The final symbolic expression is obtained by extracting the expression tree induced by the remaining active connections in the CEQL architecture.

\subsection{Evaluation of Discovered Symbolic Models}
Symbolic regression aims to fit observed data with an explicit formula that is both accurate and compact. Many existing works report symbolic criteria such as exact match or algebraic equivalence. In practice, however, these criteria are brittle: algebraic simplification is not always reliable, and many distinct expressions can fit the data equally well while remaining non-equivalent or difficult to compare symbolically. This concern is consistent with recent benchmark studies showing that standard SR datasets and evaluation protocols can overestimate scientific-discovery performance when they rely only on simplified sampling regimes or coarse symbolic metrics \cite{matsubara2024rethinking}.

In this work, we primarily evaluate models using mean squared error (MSE) on both an interpolation test set (within the training domain) and an extrapolation test set (outside the training domain). While interpolation assesses how well a model fits data within the observed range, extrapolation evaluates its ability to generalize beyond it. Specifically, the extrapolation set is constructed by sampling inputs from regions that lie strictly outside the training domain, ensuring that the model is evaluated on previously unseen value ranges.

This distinction is particularly important for symbolic regression, whose objective is to recover the underlying functional relationship rather than to interpolate between samples. In the synthetic benchmarks considered here, a compact ground-truth expression is known to exist by construction. Models that merely rely on flexible function approximation can achieve low interpolation error but typically fail to generalize outside the training domain. In contrast, models that correctly identify the underlying functional form are expected to maintain low error in extrapolation. Consequently, low extrapolation error serves as a practical proxy for recovering the true governing expression up to algebraic equivalence, and provides a more stringent measure of robustness.

To assess interpretability, we additionally report expression size measured by the node count of the inferred expression tree. These metrics capture both predictive accuracy and the simplicity of the recovered expression.

\section{Results}
This section evaluates the performance of the proposed CEQL method against representative symbolic regression baselines from three major methodological families: the GP-based PySR algorithm, the linear sparse regression method SINDy, and the non-linear EQL variant with division. In addition, we demonstrate the  capabilities on a real-world task by approximating the frequency response function of a cantilever steel beam.

\subsection{Symbolic Regression Benchmarks} \label{sec:sr_benchmarks}
To evaluate performance of the proposed method across increasing structural complexity, presence of singularities or domain restrictions, we construct a controlled benchmark set of analytic expressions. The benchmark expressions, summarized in Table~\ref{tab:sr_benchmarks}, range from simple linear and polynomial forms to expressions involving rational functions, logarithms, and square roots.

\begin{table}[t]
    \centering
    \small
    \caption{The set of SR benchmark expressions.}
    \label{tab:sr_benchmarks}
    \setlength{\tabcolsep}{4pt}
    \begin{tabular}{cll|cll}
        \toprule
        \# & Expression & Ill-posedness & \# & Expression & Ill-posedness \\
        \midrule
        E-1  & $1.87\,x_1 + 2.01$ & -- & E-6  & $2.31\,\sqrt{2.52\,x_1^2 - 1.52\,x_1 - 2.24}$ & undef. \\
        E-2  & $1.56\,x_1 + 1.59\,x_2 - 2.91$ & -- & E-7  & $\dfrac{0.53 - 2.94\,x_1}{2.32\,x_1 + 1.80}$ & pole (1) \\
        E-3  & $2.48\,x_1^2 + 1.92\,x_1 - 0.68$ & -- & E-8  & $\dfrac{1.00\,x_1 + 2.48\,x_2 - 1.36}{2.26\,x_1 - 0.91\,x_2 + 1.94}$ & pole (1) \\
        E-4  & $0.55\,x_1^2 + 2.45\,x_1 x_2 + 2.95\,x_2^2$ & -- & E-9  & $\dfrac{2.84\,x_1^2 + 1.84\,x_1 - 2.33}{-0.66\,x_1^2 + 2.94\,x_1 + 1.35}$ & pole (1) \\
             & $\quad + 1.65\,x_1 + 0.80\,x_2 + 0.86$ &  \\
        E-5  & $-2.05\,\log(1.56\,x_1^2 - 0.55\,x_1 - 2.15)$ & undef. & E-10 & $\dfrac{-1.08\,x_1^2 - 2.85\,x_1 - 2.08}{2.56\,x_1^2 + 1.78\,x_1 - 0.74}$ & pole (2) \\
        \bottomrule
    \end{tabular}
\end{table}

The expressions are generated symbolically and evaluated exactly. Coefficients are sampled independently from a signed uniform distribution $c = s \cdot u$, with $s \in \{-1,+1\}$ and $u \sim \mathcal{U}(0.5, 3.0)$, and rounded to two decimal places. For each expression, three separate datasets are generated: a training set, an interpolation test set sampled from the same domain, and an extrapolation test set sampled strictly outside the training domain. 

We evaluate symbolic models discovered by CEQL in both interpolation and extrapolation regimes. Training and interpolation samples are drawn from $[-2,2]^d$ , whereas extrapolation samples are drawn from the disjoint outer region  $([-4,-2] \cup [2,4])^d$, ensuring evaluation on previously unseen regions of the input space. The datasets contain $128$ training points and $8192$ points for both interpolation and extrapolation testing. Samples producing non-finite, undefined, or values exceeding $10^2$ are discarded. This evaluation setup is particularly relevant for assessing SR models under assumption that the underlying governing model has a compact symbolic form. SR models that identify the correct functional form are expected to generalize consistently across domains, including regions outside the training range.

Beyond polynomials, the benchmark expressions are constructed to expose structural challenges for gradient-based symbolic regression methods. Rational expressions contain denominators with one or more poles inside the training domain, while explicitly excluding poles from the extrapolation domain. Logarithmic and square-root expressions use quadratic arguments that change sign within the training domain, creating intervals where the target function is undefined. In real-valued formulations, such structures induce ill-conditioned loss landscapes, and invalid forward evaluations.

\begin{table*}[t]
    \centering
    \small
    \caption{Experimental results of SR methods applied to expressions with singularities and domain constraints.}
    \label{tab:sr_results}
    \setlength{\tabcolsep}{10pt}
    \begin{tabular}{llllll}
        \toprule
        Expression & Metric & PySR & SINDy & $\mathrm{EQL}_{\div}$ & CEQL (our) \\
        \midrule

        \multirow{3}{*}{E-1}
          & Interp. MSE & $1.6\times 10^{-15}$ & $0.0$                & $8.2\times 10^{-7}$ & $1.0\times 10^{-12}$ \\
          & Extrap. MSE & $2.4\times 10^{-15}$ & $0.0$                & $5.3\times 10^{-3}$ & $3.6\times 10^{-12}$ \\
          & NC          & $6.8$                & $5.0$                & $37.0$              & $5.0$ \\
        \midrule

        \multirow{3}{*}{E-2}
          & Interp. MSE & $6.0\times 10^{-15}$ & $5.1\times 10^{-30}$ & $5.7\times 10^{-3}$ & $2.7\times 10^{-12}$ \\
          & Extrap. MSE & $1.7\times 10^{-14}$ & $1.5\times 10^{-29}$ & $2.9\times 10^{-1}$ & $7.5\times 10^{-12}$ \\
          & NC          & $8.0$                & $8.0$                & $115.4$             & $8.0$ \\
        \midrule

        \multirow{3}{*}{E-3}
          & Interp. MSE & $4.6\times 10^{-15}$ & $1.5\times 10^{-30}$ & $1.1\times 10^{-1}$ & $6.8\times 10^{-11}$ \\
          & Extrap. MSE & $2.7\times 10^{-14}$ & $6.9\times 10^{-29}$ & $1.3\times 10^{2}$  & $2.0\times 10^{-9}$ \\
          & NC          & $13.0$               & $10.0$               & $43.0$              & $10.0$ \\
        \midrule

        \multirow{3}{*}{E-4}
          & Interp. MSE & $5.8\times 10^{-9}$  & $5.7\times 10^{-29}$ & $6.2\times 10^{-1}$ & $9.0\times 10^{-11}$ \\
          & Extrap. MSE & $2.4\times 10^{-7}$  & $4.9\times 10^{-28}$ & $4.2\times 10^{5}$  & $2.4\times 10^{-9}$ \\
          & NC          & $19.6$               & $22.0$               & $151.0$             & $22.0$ \\
        \midrule

        \multirow{3}{*}{E-5}
          & Interp. MSE & $6.4\times 10^{-5}$  & $1.1$                & $1.9\times 10^{0}$  & $5.9\times 10^{-9}$ \\
          & Extrap. MSE & $3.6\times 10^{-7}$  & $1.7\times 10^{3}$   & $7.5\times 10^{2}$  & $3.4\times 10^{-11}$ \\
          & NC          & $20.2$               & $19.0$               & $43.0$              & $18.4$ \\
        \midrule

        \multirow{3}{*}{E-6}
          & Interp. MSE & $9.1\times 10^{-14}$ & $1.3\times 10^{-1}$  & $4.7\times 10^{-1}$ & $3.8\times 10^{-7}$ \\
          & Extrap. MSE & $1.0\times 10^{-13}$ & $1.3\times 10^{2}$   & $3.6\times 10^{1}$  & $2.1\times 10^{-9}$ \\
          & NC          & $15.0$               & $19.0$               & $43.0$              & $21.4$ \\
        \midrule

        \multirow{3}{*}{E-7}
          & Interp. MSE & $4.1\times 10^{-5}$  & $5.0\times 10^{1}$   & $1.3\times 10^{3}$  & $8.6\times 10^{-8}$ \\
          & Extrap. MSE & $1.7\times 10^{-12}$ & $6.5\times 10^{0}$   & $1.8\times 10^{3}$  & $1.9\times 10^{-9}$ \\
          & NC          & $15.0$               & $19.0$               & $43.0$              & $26.8$ \\
        \midrule

        \multirow{3}{*}{E-8}
          & Interp. MSE & $2.9\times 10^{-4}$  & $7.1\times 10^{1}$   & $1.3\times 10^{3}$  & $9.3\times 10^{-8}$ \\
          & Extrap. MSE & $2.5\times 10^{-4}$  & $2.1\times 10^{2}$   & $7.1\times 10^{4}$  & $1.5\times 10^{-9}$ \\
          & NC          & $16.8$               & $46.0$               & $151.0$             & $33.2$ \\
        \midrule

        \multirow{3}{*}{E-9}
          & Interp. MSE & $5.1\times 10^{-5}$  & $3.1\times 10^{1}$   & $1.3\times 10^{2}$  & $4.8\times 10^{-8}$ \\
          & Extrap. MSE & $2.1\times 10^{0}$   & $3.3\times 10^{1}$   & $8.4\times 10^{1}$  & $9.7\times 10^{-7}$ \\
          & NC          & $20.6$               & $19.0$               & $43.0$              & $37.2$ \\
        \midrule

        \multirow{3}{*}{E-10}
          & Interp. MSE & $8.0\times 10^{-8}$  & $4.6\times 10^{1}$   & $7.0\times 10^{2}$  & $2.7\times 10^{-6}$ \\
          & Extrap. MSE & $6.4\times 10^{-13}$ & $4.0\times 10^{1}$   & $7.9\times 10^{0}$  & $4.2\times 10^{-11}$ \\
          & NC          & $18.6$               & $19.0$               & $43.0$              & $29.4$ \\
        \bottomrule
    \end{tabular}
\end{table*}

Table~\ref{tab:sr_results} summarizes the experimental results of different SR methods applied to the generated benchmark datasets. Complete evaluation results are reported in Appendix~\ref{si:benchmark_results}. We compare the proposed CEQL method against SINDy, EQL$_{\div}$, and PySR, a state-of-the-art GP-based SR algorithm. The table reports the mean squared error (MSE) on the interpolation and extrapolation test sets, as well as the node count (NC) of the recovered expressions, across five independent runs. The hyperparameter settings for all methods are reported in Appendix~\ref{si:benchmark_hyperparameters}.

While SINDy demonstrates extremely low test errors for polynomial expressions, it fails to correctly determine nonlinear terms (E-5 -- E-10), including rational expressions. This occurs because the SINDy library does not contain the required terms. The coefficients inside nonlinear terms in SINDy are not learnable, and covering all possible non-linear terms with the correct internal coefficients is computationally infeasible, revealing a fundamental limitation of linear SR algorithms. EQL$_{\div}$ as well fails to achieve low approximation error on nonlinear benchmarks. Although EQL$_{\div}$ includes division operator as an output node, the algorithm is unable to learn rational structures in the presence of poles within the training domain.

PySR and CEQL approaches achieve in general low errors on the benchmark expressions. While PySR reaches lower extrapolation errors in comparison to CEQL, performance of PySR on the interpolation test set degrades faster as complexity of expression increases. For simpler expressions, such as E-1--E-4 and E-6, the zero variance in the node count (NC) indicates that CEQL consistently converges to the same sparsified symbolic structure. For more complex rational expressions, the non-zero variance in NC suggests that additional terms with negligible impact on MSE may persist in some runs. This behavior is likely due to the larger range of the target values in rational expressions, which reduces the sensitivity of the loss to small residual terms. Although determining strict algebraic equivalence between the recovered and ground-truth expressions is nontrivial, the comparable magnitudes of interpolation and extrapolation errors provide indirect evidence that the underlying functional form is recovered. The recovered symbolic expressions with the lowest training MSE across five independent CEQL runs are provided in Appendix~\ref{si:predicted_expressions}, where the reader can try to manually verify whether algebraic equivalence is preserved.

\subsection{Frequency Response Function Discovery}

This experiment is designed to evaluate if CEQL can recover pole-like structure from real measured data, rather than only from synthetic benchmarks. Frequency response functions of lightly damped structures exhibit sharp resonance peaks whose locations shift with changes in attached masses; these peaks are naturally explained by rational forms with denominators that approach zero near modal frequencies. The goal is therefore to learn an interpretable symbolic surrogate of the measured response magnitude that can represent both the broadband trend and the resonance spikes, and whose denominators can be inspected to understand how resonance frequencies shift as a function of structural damage.

The dataset consists of repeated experimental measurements of the frequency response of a cantilever steel beam equipped with six detachable masses \cite{de2024experimental}. Each experiment yields a complex-valued inertance response over $f \in [0,2000]$ Hz, recorded as magnitude $y$ (in dB) and phase $\phi$ (in degrees). Four structural conditions are included: a healthy configuration with all masses attached and three progressively damaged configurations created by removing a subset of masses. Across repetitions, the remaining masses are placed at randomly perturbed longitudinal positions along the beam. In our setup, we learn a symbolic regressor for the response magnitude as a function of the excitation frequency and the structural condition. The input variables are the excitation frequency $\omega$ (in kHz) and a scalar damage indicator $d$, representing the percentage of the mass loss for each experiment. The target variable is the magnitude of the inertance response in linear scale.

Let $\omega$ denote the excitation frequency (in kHz) and $d$ the damage level expressed as the percentage of removed masses. The training data therefore consist of tuples $(\omega, d, y)$, where $y$ is the linear-scale magnitude of the measured frequency response function.

The model used in this experiment is a symbolic architecture designed to represent the structure of frequency response functions. The predicted response magnitude is modeled as a linear broadband trend combined with a sum of resonant terms

\begin{equation} \label{eq:frf_structure}
    \hat{H}(\omega,d)
    = a\omega + b
    + \sum_{i=1}^{n}
    \frac{A_i}
    {\left(\omega - h_i(d)\right)^2 + \gamma_i},
\end{equation}

where $a$ and $b$ represent the broadband trend, $A_i$ and $\gamma_i$ are learned coefficients controlling the amplitude and width of the $i$-th resonance, and $h_i(d)$ denotes a symbolic function describing how the corresponding resonance frequency shifts with damage level $d$. The model is initialized with a fixed number of resonant components $n=20$. During training, periodic pruning of the internal CEQL weights removes connecting edges, effectively reducing the number of active resonant terms. The minimum number of edges is set to 50.

Each function $h_i(d)$ is generated by a CEQL symbolic network that receives the damage variable $d$ and produces a nonlinear expression representing the damage-dependent resonance location. The CEQL component consists of two symbolic layers with a fixed operator library. Each layer uses the operators $\{\mathrm{id},\mathrm{const},\mathrm{square},\mathrm{mul}\}$. Through these layers the model learns compact symbolic expressions for the resonance shifts $h_i(d)$. The resulting architecture therefore balances interpretability and structural simplicity with the expressive capacity required to approximate the measured frequency response.

\begin{figure}
    \centering
    \begin{subfigure}{0.4\textwidth}
        \centering
        \includegraphics[width=\linewidth]{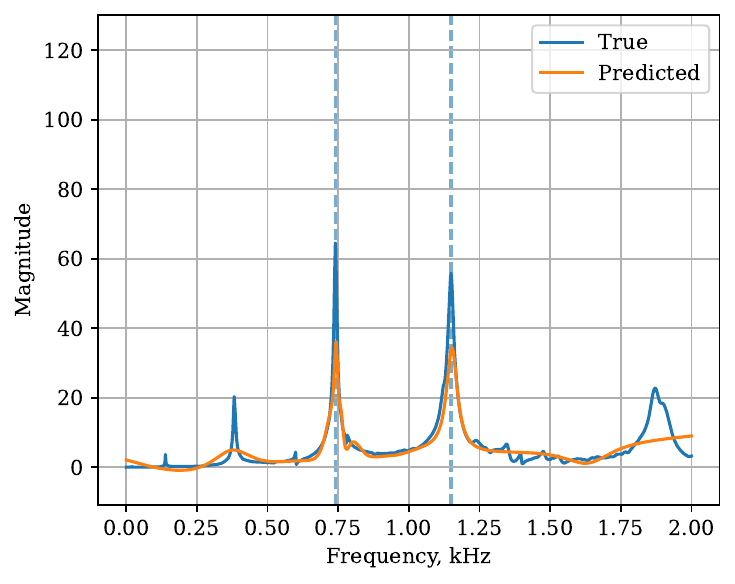}
        \caption{Healthy beam}
    \end{subfigure}
    \begin{subfigure}{0.4\textwidth}
        \centering
        \includegraphics[width=\linewidth]{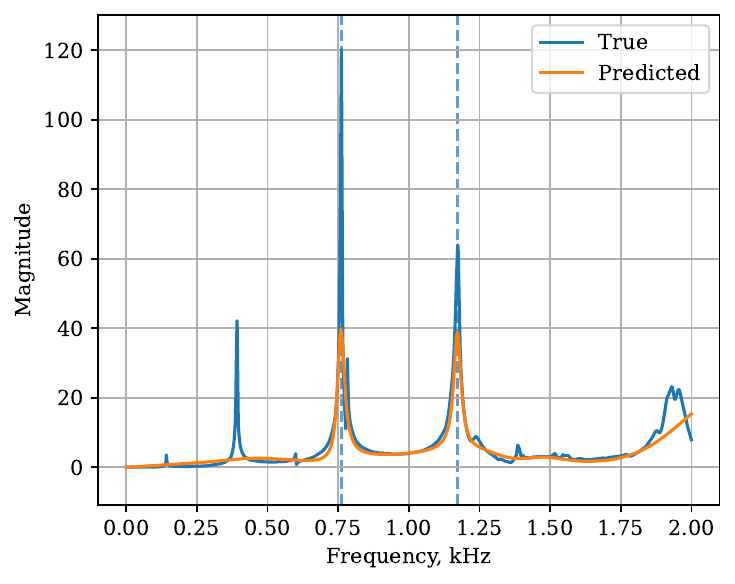}
        \caption{Damaged beam (5.92\%)}
    \end{subfigure}
    \caption{Measured and predicted byt the CEQL model FRF magnitudes for (a) a healthy beam and (b) a beam with $5.92\%$ damage. Dashed vertical lines indicate the locations of two dominant resonance peaks detected from the measured response.}
    \label{fig:frf_prediction}
\end{figure}

Figure~\ref{fig:frf_prediction} shows the predicted frequency response functions together with the measured responses for healthy and damaged beam configurations. The symbolic model captures the overall structure of the FRF, including the broadband trend and the dominant resonance peaks. In particular, the model reproduces the two major resonances around $0.75$~kHz and $1.18$~kHz and correctly captures their shifts across damage levels. The predicted peaks appear smoother and slightly lower in amplitude than the measured responses. This behavior is consistent with the structure of the model in Equation~\ref{eq:frf_structure}, where the damage dependence is introduced only through the resonance location $h_i(d)$, while the numerator coefficients $A_i$ and the denominator constants $\gamma_i$ remain fixed for all damage levels. As a result, the model has sufficient flexibility to adjust the position of each resonance but limited freedom to adapt the exact peak shape or height across damage conditions. Consequently, the symbolic model prioritizes accurate prediction of resonance locations while providing a simplified approximation of the peak amplitudes.

The symbolic expression obtained after training and pruning is reported in Appendix~\ref{si:frf_expression}. The resulting model contains several resonant terms whose denominators follow the structure of Equation~\ref{eq:frf_structure}, where the resonance locations are expressed as nonlinear functions of the damage variable $d$. This form allows the model to explicitly describe how modal frequencies shift as the structural condition changes. Several rational terms correspond to resonances located near the dominant peaks observed in the measured frequency response, while the linear terms contribute to approximating the broadband behavior of the FRF. Due to the pruning procedure, the final expression is substantially simpler than the initial architecture, retaining only the resonant components necessary to approximate the measured response while satisfying the minimal number of edges condition.

\begin{figure}
    \centering
    \begin{subfigure}{0.45\textwidth}
        \centering
        \includegraphics[width=\linewidth]{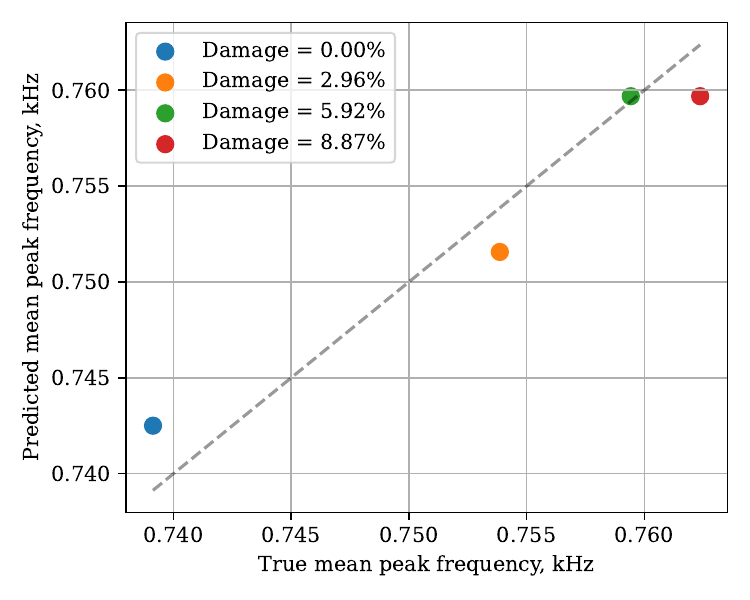}
        \caption{Peak 1 comparison}
    \end{subfigure}
    \begin{subfigure}{0.45\textwidth}
        \centering
        \includegraphics[width=\linewidth]{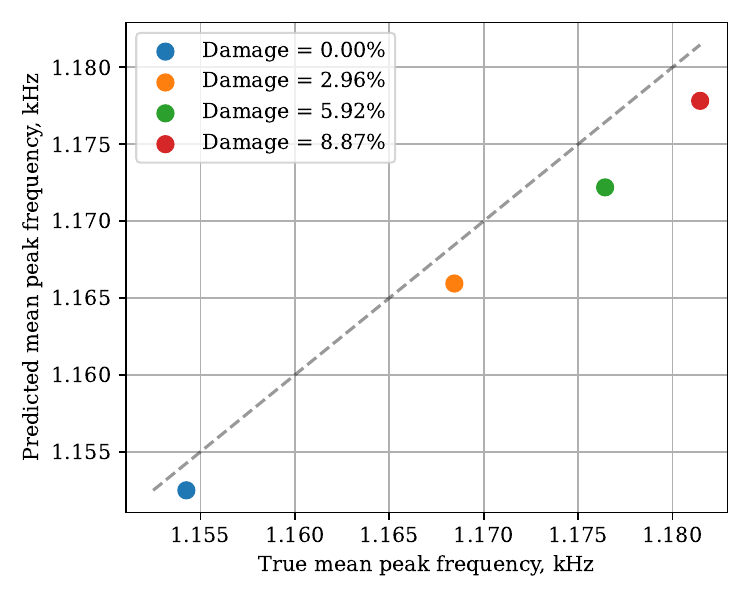}
        \caption{Peak 2 comparison}
    \end{subfigure}
    \caption{Comparison of predicted and measured resonance frequencies for the first (a) and second (b) peaks across all damage levels. Each point corresponds to the mean peak frequency computed from multiple FRF measurements.}
    \label{fig:frf_peaks}
\end{figure}

To quantify how well the symbolic model captures the resonance shifts, Figure~\ref{fig:frf_peaks} compares the predicted and measured mean frequencies of the two main peaks across all damage levels. For each structural condition, the peak frequency is estimated within a fixed window around the expected resonance location and averaged over repeated FRF measurements. The predicted peak locations closely follow the measured values and lie near the identity line, indicating that the model accurately captures the damage-dependent shift of resonance frequencies. Small deviations appear mainly for the healthy configuration and for higher damage levels, but the overall trend across damage states is preserved for both resonances.

\section{Discussion}

The results highlight that the primary difficulty of learning rational expressions in EQL-type architectures is not representational capacity but optimization. In particular, the instability observed when learning divisions appears to stem from real-valued degeneracies around zero-crossings in the denominator, where gradient contributions with opposite signs cancel. Introducing complex-valued weights changes the optimization geometry, enabling gradient-based training to bypass these degeneracies and converge to rational structures that remain well defined on the real axis.

From an application perspective, this behavior is especially relevant in engineering settings, where rational dependencies and poles are often fundamental. Recovering poles explicitly as symbolic denominators improves interpretability and enables direct reasoning about resonance behavior and stability margins.

The proposed method also retains a structural property of EQL-style models that may be advantageous in higher-dimensional settings. Rather than performing an explicit combinatorial search over expression trees, CEQL optimizes the weights of a fixed differentiable architecture defined by a chosen operator library. This shifts the symbolic regression problem from discrete structure search to continuous parameter optimization. In principle, such a formulation may scale more favorably with input dimensionality than genetic-programming-based approaches, whose search spaces typically expand rapidly as the number of variables and candidate compositions increases.

At the same time, the results expose clear limitations of gradient-based symbolic learning that are not resolved by the proposed extension. Trigonometric operators remain particularly challenging to identify over wide input domains. The oscillatory nature of  $\sin(\cdot)$ and $\cos(\cdot)$ leads to  frequent sign changes in per-sample gradients, which in turn causes substantial  cancellation when gradients averaged across data. As a consequence, optimization provides weak or misleading signals for frequency and phase parameters unless initialization is already close to the target solution. This observation is consistent with prior reports that successful recovery of trigonometric expressions in EQL-based models often relies on carefully chosen initializations, and it suggests that architectural or optimization-level modifications are required to address this limitation more fundamentally. 

A related challenge concerns sparsity enforcement. While smooth sparsity penalties and pruning strategies are effective in reducing expression size, the results indicate that their interaction with optimization remains nontrivial. Aggressive regularization or early pruning can irreversibly remove correct components of an expression, whereas insufficient regularization leaves residual terms that obscure the underlying structure. This sensitivity suggests that more principled sparsity control mechanisms, potentially informed by uncertainty estimates, stability criteria, or post hoc symbolic simplification, are necessary for consistent recovery of minimal expressions.

While these findings suggest that extending symbolic regression models with complex-valued optimization alleviate specific degeneracies associated with rational expressions, other structural classes, most notably periodic functions, remain fundamentally misaligned with standard gradient aggregation over wide domains. Similarly, sparsity control emerges as an optimization-sensitive design choice rather than a purely regularization-driven one.

\section{Conclusion}
This work introduced Complex Equation Learner (CEQL), an extension of the Equation Learner framework that uses complex weights together with a tailored optimization strategy to expand the class of symbolic operators that can be learned reliably with gradient-based methods. The primary challenge addressed is the failure of real-valued optimization in the presence of sign-changing gradients and near-singular structures, most prominently those arising from division. By optimizing in the complex domain and projecting predictions back onto the real axis, CEQL mitigates gradient-cancellation pathologies that prevent standard EQL variants from discovering rational expressions with sign-changing denominators and singular behavior.

Across synthetic benchmarks with singularity-inducing operators, CEQL trains stably and recovers compact analytical expressions in regimes where real-valued baselines become unreliable. On the benchmark tasks considered (Table~\ref{tab:sr_results}), CEQL exhibits consistent recovery of symbolic models that demonstrate low prediction error in both interpolation and extrapolation regimes, indicating that the learned expressions capture the underlying functional structure rather than overfitting sampled training points. In addition, the cantilever-beam frequency response experiment shows that CEQL can recover rational near-pole structure from real measured data, producing interpretable expressions that support direct analysis of damage-dependent resonance shifts.

Several directions for future work remain. Learning periodic trigonometric structures continues to be challenging due to periodic sign changes that lead to weak or canceling gradient signals over large domains, suggesting the need for surrogate operators or continuation-based training schemes to robustly recover  frequency and phase parameters. Enforcing sparsity in a stable and reproducible manner also remains difficult;  improved pruning criteria and regularization schedules are required to consistently obtain minimal symbolic expressions. More broadly, the complex-parameter perspective provides a general mechanism for reshaping ill-conditioned real-valued loss landscapes and may prove useful for other non-smooth  operators beyond symbolic regression. Finally, because CEQL outputs explicit rational forms, it is naturally aligned with rational approximation and pole-matching tasks, where interpretable denominators are primary objects of interest.

\bibliographystyle{plainnat}
\bibliography{main}

\appendix
\section{Illustration of Complex-Domain Optimization} \label{si:complex_optimization}

\begin{figure}[htbp]
    \centering
    \includegraphics[width=0.45\textwidth]{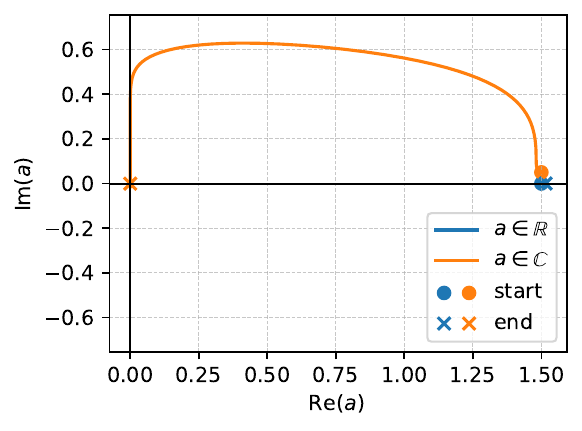}
    \caption{Optimization trajectories of the parameter $a$ in the complex plane for the task of approximating $f(x)=1/x$ with $\hat{f}(x)=1/(x+a)$.}
    \label{fig:division_convergence}
\end{figure}

To illustrate the effect of complex-valued optimization for division, we consider approximating $f(x)=1/x$ with the model $\hat f(x)=1/(x+a)$. Figure~\ref{fig:division_convergence} compares the trajectory of the parameter $a$ under Adam when $a$ is constrained to $\mathbb{R}$ and when $a$ is allowed to vary in $\mathbb{C}$. The target function is sampled at 100 points uniformly from $[-3,3]$, and the loss is defined as the $\ell_2$ approximation error.

\FloatBarrier

\section{Hyperparameters and Experimental Settings} \label{si:benchmark_hyperparameters}
This section reports all hyperparameters used for the symbolic regression benchmarks E-1--E-10. Unless explicitly stated, all hyperparameters are fixed across benchmark expressions and across independent runs.

\subsection{Complex Equation Learner (CEQL)}

\begin{table}[htbp]
\centering
\small
\caption{CEQL hyperparameters used for benchmarks E-1--E-10.}
\label{si:ceql_hyperparameters}
    \begin{tabular}{ll}
        \toprule
        \textbf{Component} & \textbf{Setting} \\
        \midrule
        Optimizer & Adam \\
        Learning rate (LR) & $10^{-2}$ \\
        LR scheduler (phase 3) & ReduceLROnPlateau \\
        LR scheduler patience & $2000$ \\
        LR scheduler factor & $0.1$ \\
        Minimum LR & $10^{-5}$ \\
        Convergence threshold & $10^{-7}$ \\
        Loss & Pivoted Relative Mean Squared Error (PR-MSE) \\
        Epochs & $100{,}000$ (phase 1), $200{,}000$ (phase 2), $50{,}000$ (phase 3) \\
        Im($w$) lambda & $10^{-10}$ (Phase 1), $10^{-3}$ (Phase 2), $10^{3}$ (Phase 3) \\
        $\| w \|_1$ lambda & $10^{-10}$ (Phase 1), $10^{-7}$ (Phase 2), $10^{-7}$ (Phase 3) \\
        Log, sqrt argument penalty & $10^{-10}$ (Phase 1), $10^{3}$ (Phase 2), $10^{3}$ (Phase 3) \\
        Symbolic layers & $2$ \\
        Library operators (per layer) & ($const \times 2$, $(.)^2 \times 2$, $const \times 2$, $mul \times 2$), \\
         & ($id$, $\log{.} \times 2$, $\sqrt{.} \times 2$, $\div \times 2$) \\
        Pruning enabled & Phase 2 \\
        Pruning interval, epochs & $10{,}000$ \\
        Pruning fraction & $0.1$ \\
        Minimal number of edges & $15$ \\
        \bottomrule
    \end{tabular}
\end{table}

\FloatBarrier

\subsection{PySR}

\begin{table}[htbp]
\centering
\small
\caption{PySR hyperparameters used for benchmarks E-1--E-10.}
\label{si:pysr_hyperparameters}
    \begin{tabular}{ll}
        \toprule
        \textbf{Component} & \textbf{Setting} \\
        \midrule
        Number of iterations & 1000 \\
        Population size & 27 \\
        Number of parallel populations & 200 \\
        Max sequence size & 20 \\
        Unary operators & ($\log$, $\sqrt{.}$, $.^2$) \\
        Binary operators & ($+$, $-$, $\times$, $\div$) \\
        Element-wise loss & $(x-y)^2$ \\
        Model selection & best \\
        Crossover probability & $0.0259$ \\
        Tournament selection n & 15 \\
        Tournament selection p & $0.982$ \\
        \bottomrule
    \end{tabular}
\end{table}

\FloatBarrier

\subsection{EQL with division}

\begin{table}[htbp]
\centering
\small
\caption{EQL$_{\div}$ hyperparameters used for benchmarks E-1--E-10.}
\label{si:eqldiv_hyperparameters}
    \begin{tabular}{ll}
        \toprule
        \textbf{Component} & \textbf{Setting} \\
        \midrule
        Optimizer & SGD \\
        Learning rate & $10^{-2}$ \\
        Number of epochs & $10{,}000$ \\
        Batch size & 64 \\
        Number of symbolic layers & 2 \\
        Units per base function & 3 \\
        Unary operators & $(id)$ \\
        Binary operators & $(mul)$ \\
        Output node & $\div$ \\
        $\|w\|_1$ regularization & $10^{-3}$ \\
        $\|w\|_2$ regularization & $0$ \\
        Regularization start epoch & 500 \\
        Regularization end epoch & 9500 \\
        Validation interval & 100 epochs \\
        Division regularization parameter $k$ & 50 \\
        Symbolic pruning threshold & $10^{-2}$ \\
        Active-unit threshold & $10^{-2}$ \\
        \bottomrule
    \end{tabular}
\end{table}

\FloatBarrier

\subsection{SINDy}

\begin{table}[htbp]
    \centering
    \small
    \caption{SINDy hyperparameters used for benchmarks E-1--E-10.}
    \label{si:sindy_hyperparameters}
    \begin{tabular}{ll}
        \toprule
        \textbf{Component} & \textbf{Setting} \\
        \midrule
        Polynomial degree & 2 \\
        Include interactions & True \\
        Include polynomial bias term & True \\
        Unary operators & ($\log$, $\sqrt{.}$) \\
        Binary operators & ($\div$) \\
        Sparse optimizer & STLSQ \\
        Sparsity threshold & $10^{-2}$ \\
        Ridge regularization $\alpha$ & $10^{-6}$ \\
        Maximum optimizer iterations & 1000 \\
        Normalize library columns & True \\
        Log stabilization $\epsilon$ & $10^{-12}$ \\
        Division stabilization $\epsilon$ & $10^{-6}$ \\
        Use absolute value in $\sqrt{\cdot}$ & True \\
        Maximum feature magnitude (clipping) & $10^{6}$ \\
        Coefficient zero tolerance & $10^{-12}$ \\
        Number of runs per benchmark & 1 \\
        \bottomrule
    \end{tabular}
\end{table}

\FloatBarrier

\section{Benchmark Results} \label{si:benchmark_results}

\begin{table*}[htbp]
    \centering
    \small
    \caption{Experimental results of SR methods applied to expressions with singularities and domain constraints. The table reports the mean value of the metric and its standard deviation obtained across 5 independent runs.}
    \label{tab:full_benchmark_results}
    \setlength{\tabcolsep}{2pt}
    \begin{tabular}{l ccc ccc}
        \toprule
        & \multicolumn{3}{c}{PySR} & \multicolumn{3}{c}{SINDy} \\
        \cmidrule(lr){2-4} \cmidrule(lr){5-7}
        \# & Interp. MSE & Extrap. MSE & NC & Interp. MSE & Extrap. MSE & NC \\
        \midrule
        E-1  & $(1.6\pm2.4)\times 10^{-15}$ & $(2.4\pm3.3)\times 10^{-15}$ & $6.8\pm3.6$ & $0.0$ & $0.0$ & $5.0$ \\
        E-2  & $(6.0\pm7.3)\times 10^{-15}$ & $(1.7\pm2.1)\times 10^{-14}$ & $8.0\pm0.0$ & $5.1\times 10^{-30}$ & $1.5\times 10^{-29}$ & $8.0$ \\
        E-3  & $(4.6\pm4.7)\times 10^{-15}$ & $(2.7\pm3.2)\times 10^{-14}$ & $13.0\pm1.1$ & $1.5\times 10^{-30}$ & $6.9\times 10^{-29}$ & $10.0$ \\
        E-4  & $(5.8\pm5.5)\times 10^{-9}$ & $(2.4\pm2.7)\times 10^{-7}$ & $19.6\pm0.5$ & $5.7\times 10^{-29}$ & $4.9\times 10^{-28}$ & $22.0$ \\
        E-5  & $(6.4\pm10.3)\times 10^{-5}$ & $(3.6\pm3.2)\times 10^{-7}$ & $20.2\pm5.7$ & $1.1$ & $1.7\times 10^{3}$ & $19.0$ \\
        E-6  & $(9.1\pm11.9)\times 10^{-14}$ & $(1.0\pm0.8)\times 10^{-13}$ & $15.0\pm3.3$ & $1.3\times 10^{-1}$ & $1.3\times 10^{2}$ & $19.0$ \\
        E-7  & $(4.1\pm8.1)\times 10^{-5}$ & $(1.7\pm3.1)\times 10^{-12}$ & $15.0\pm3.3$ & $5.0\times 10^{1}$ & $6.5\times 10^{0}$ & $19.0$ \\
        E-8  & $(2.9\pm2.9)\times 10^{-4}$ & $(2.5\pm3.5)\times 10^{-4}$ & $16.8\pm1.0$ & $7.1\times 10^{1}$ & $2.1\times 10^{2}$ & $46.0$ \\
        E-9  & $(5.1\pm6.6)\times 10^{-5}$ & $(2.1\pm2.6)\times 10^{0}$ & $20.6\pm1.0$ & $3.1\times 10^{1}$ & $3.3\times 10^{1}$ & $19.0$ \\
        E-10 & $(8.0\pm10.3)\times 10^{-8}$ & $(6.4\pm6.7)\times 10^{-13}$ & $18.6\pm2.2$ & $4.6\times 10^{1}$ & $4.0\times 10^{1}$ & $19.0$ \\
        \midrule
        & \multicolumn{3}{c}{CEQL (our)} & \multicolumn{3}{c}{$\mathrm{EQL}_{\div}$} \\
        \cmidrule(lr){2-4} \cmidrule(lr){5-7}
        \# & Interp. MSE & Extrap. MSE & NC & Interp. MSE & Extrap. MSE & NC \\
        \midrule
        E-1  & $(1.0\pm0.8)\times 10^{-12}$ & $(3.6\pm4.5)\times 10^{-12}$ & $5.0\pm0.0$   & $(8.2\pm6.2)\times 10^{-7}$ & $(5.3\pm8.0)\times 10^{-3}$ & $37.0 \pm 5.8$ \\
        E-2  & $(2.7\pm1.5)\times 10^{-12}$ & $(7.5\pm3.9)\times 10^{-12}$ & $8.0\pm0.0$   & $(5.7\pm6.4)\times 10^{-3}$ & $(2.9\pm5.4)\times 10^{-1}$ & $115.4 \pm 38.1$ \\
        E-3  & $(6.8\pm10.1)\times 10^{-11}$ & $(2.0\pm3.2)\times 10^{-9}$ & $10.0\pm0.0$  & $(1.1\pm0.7)\times 10^{-1}$ & $(1.3\pm0.7)\times 10^{2}$ & $43.0 \pm 0.0$ \\
        E-4  & $(9.0\pm5.8)\times 10^{-11}$ & $(2.4\pm2.1)\times 10^{-9}$  & $22.0\pm0.0$  & $(6.2\pm3.0)\times 10^{-1}$ & $(4.2\pm5.9)\times 10^{5}$ & $151.0 \pm 0.0$ \\
        E-5  & $(5.9\pm7.2)\times 10^{-9}$  & $(3.4\pm4.3)\times 10^{-11}$  & $18.4\pm4.3$ & $(1.9\pm1.1)\times 10^{0}$ & $(7.5\pm14.9)\times 10^{2}$ & $43.0 \pm 0.0$ \\
        E-6  & $(3.8\pm4.7)\times 10^{-7}$ & $(2.1\pm3.7)\times 10^{-9}$ & $21.4\pm9.3$  & $(4.7\pm9.1)\times 10^{-1}$ & $(3.6\pm3.8)\times 10^{1}$ & $43.0 \pm 0.0$ \\
        E-7  & $(8.6\pm7.7)\times 10^{-8}$  & $(1.9\pm3.7)\times 10^{-9}$  & $26.8\pm12.4$ & $(1.3\pm2.3)\times 10^{3}$ & $(1.8\pm3.6)\times 10^{3}$ & $43.0 \pm 0.0$ \\
        E-8  & $(9.3\pm14.2)\times 10^{-8}$ & $(1.5\pm2.4)\times 10^{-9}$  & $33.2\pm8.4$  & $(1.3\pm2.2)\times 10^{3}$ & $(7.1\pm14.0)\times 10^{4}$ & $151.0 \pm 0.0$ \\
        E-9  & $(4.8\pm3.5)\times 10^{-8}$   & $(9.7\pm19.4)\times 10^{-7}$   & $37.2\pm9.6$  & $(1.3\pm1.0)\times 10^{2}$ & $(8.4\pm8.5)\times 10^{1}$ & $43.0 \pm 0.0$ \\
        E-10 & $(2.7\pm1.1)\times 10^{-6}$   & $(4.2\pm6.7)\times 10^{-11}$  & $29.4\pm3.2$ & $(7.0\pm9.2)\times 10^{2}$ & $(7.9\pm13.4)\times 10^{0}$ & $43.0 \pm 0.0$ \\
        \bottomrule
    \end{tabular}
\end{table*}

\FloatBarrier

\section{Recovered Symbolic Expressions} \label{si:predicted_expressions}

\begin{table*}[htbp]
\centering
\small
\caption{
Symbolic expressions recovered by CEQL on benchmarks E-1--E-10.
For each benchmark, we report the expression obtained in the run with the lowest training MSE across five independent initializations.
}
\label{tab:predicted_expressions}
\setlength{\tabcolsep}{6pt}
    \begin{tabular}{c p{7.5cm} p{7.5cm}}
    \toprule
    ID & Target expression & Discovered expression \\
    \midrule
    E-1 & $1.87\,x_1 + 2.01$ & $1.87 x_{1} + 2.01$ \\[2mm]
    
    E-2 & $1.56\,x_1 + 1.59\,x_2 - 2.91$ & $1.56 x_{1} + 1.59 x_{2} - 2.91$ \\[2mm]
    
    E-3 & $2.48\,x_1^2 + 1.92\,x_1 - 0.68$ & $2.48 x_{1}^{2} + 1.92 x_{1} - 0.68$ \\[2mm]
    
    E-4 & $0.55\,x_1^2 + 2.45\,x_1 x_2 + 2.95\,x_2^2 + 1.65\,x_1 + 0.80\,x_2 + 0.86$ & $0.55 x_{1}^{2} + 2.45 x_{1} x_{2} + 1.64999 x_{1} + 2.95 x_{2}^{2} + 0.8 x_{2} + 0.86$ \\[4mm]
    
    E-5 & $-2.05\,\log(1.56\,x_1^2 - 0.55\,x_1 - 2.15)$ & $1.0 \cdot 10^{-5} \left(x_{1}^{2}\right)^{0.5} + 1.28808 - 2.05 \log{\left(2.92419 x_{1}^{2} - 1.03097 x_{1} - 4.03014 \right)}$ \\[2mm]
    
    E-6 & $2.31\,\sqrt{2.52\,x_1^2 - 1.52\,x_1 - 2.24}$ & $1.79564 \left(4.17046 x_{1}^{2} - 2.51552 x_{1} - 3.70708\right)^{0.5}$ \\[2mm]
    
    E-7 & $\dfrac{0.53 - 2.94\,x_1}{2.32\,x_1 + 1.80}$ & $\frac{0.79672 x_{1}}{- 1.0 x_{1} - 0.77586} - 0.47051 - \frac{0.5935}{- 1.0 x_{1} - 0.77586}$ \\[4mm]
    
    E-8 & $\dfrac{1.00\,x_1 + 2.48\,x_2 - 1.36}{2.26\,x_1 - 0.91\,x_2 + 1.94}$ & $\frac{0.59458 x_{1}}{1.0 x_{1} - 0.40265 x_{2} + 0.85841} + \frac{1.0361 x_{2}}{1.0 x_{1} - 0.40265 x_{2} + 0.85841} - 0.1521 - \frac{0.4712}{1.0 x_{1} - 0.40265 x_{2} + 0.85841}$ \\[2mm]
    
    E-9 & $\dfrac{2.84\,x_1^2 + 1.84\,x_1 - 2.33}{-0.66\,x_1^2 + 2.94\,x_1 + 1.35}$ & $0.00017 x_{1} - \frac{3.19994 x_{1}}{0.14582 x_{1}^{2} - 0.64944 x_{1} - 0.29822} - 4.30141 - \frac{0.76806}{0.14582 x_{1}^{2} - 0.64944 x_{1} - 0.29822}$ \\[2mm]
    
    E-10 & $\dfrac{-1.08\,x_1^2 - 2.85\,x_1 - 2.08}{2.56\,x_1^2 + 1.78\,x_1 - 0.74}$ & $- \frac{1.17925 x_{1}}{1.4382 x_{1}^{2} + 1.0 x_{1} - 0.41573} - 0.42187 - \frac{1.34393}{1.4382 x_{1}^{2} + 1.0 x_{1} - 0.41573}$ \\
    \bottomrule
    \end{tabular}
\end{table*}

\section{Recovered symbolic model for the FRF experiment} \label{si:frf_expression}

\begin{equation} \label{eq:frf_model}
    \begin{aligned}
        \hat{H}{\left(\omega,d \right)} &= 5.21654 \omega - 0.3154 \\
        &+ \frac{0.08445}{\left(- 0.0007 d^{4} + 0.00217 d^{3} - 0.00386 d + \omega - 0.89\right)^{2} - 0.01625} \\
        &+ \frac{0.0577}{\left(0.05572 d^{2} - 0.25854 d + \omega - 1.26425\right)^{2} - 0.1399} \\
        &+ \frac{0.47169}{\left(0.03456 d^{2} - 0.37289 d + \omega + 0.17194\right)^{2} - 0.36363} \\
        &+ \frac{2.42351}{\left(- 0.02541 d^{2} + 0.061 d + \omega - 0.23462\right)^{2} - 0.0339} \\
        &- \frac{0.76141}{\left(- 0.03146 d^{2} + 0.08937 d + \omega - 0.50138\right)^{2} + 0.04658} \\
        &- \frac{0.26376}{\left(- 0.20983 d + \omega - 0.76732\right)^{2} + 0.00551} \\
        &+ \frac{0.48817}{\left(- 0.27162 d + \omega - 0.20312\right)^{2} + 0.31504} \\
        &+ \frac{0.08136}{\left(- 0.52067 d + \omega - 0.76341\right)^{2} - 0.14816} \\
        &+ \frac{0.29455}{\left(\omega - 0.69792\right)^{2} + 0.06101},
    \end{aligned}
\end{equation}

\end{document}